\documentclass[10pt, a4paper]{article}

\usepackage[final]{lrec2026} 
\usepackage{amssymb} 
\usepackage{booktabs}
\usepackage{enumitem}
\usepackage{float}
\usepackage{xcolor}
\usepackage{amsmath} 
\newcommand{\cmark}{\ensuremath{\checkmark}}
\newcommand{\qmark}{?}
\usepackage{booktabs,makecell}
\usepackage[table]{xcolor}
\definecolor{midred}{RGB}{255,150,150}
\definecolor{midgreen}{RGB}{150,255,150}
\definecolor{midyellow}{RGB}{255,245,150}

\title{Task-Lens: Cross-Task Utility Based Speech Dataset Profiling for Low-Resource Indian Languages}

\name{Swati Sharma, Divya V. Sharma, Anubha Gupta} 

\address{SBI Lab, IIIT Delhi \\
         swati21568@iiitd.ac.in, divyas@iiitd.ac.in, anubha@iiitd.ac.in\\}

\abstract{
The rising demand for inclusive speech technologies amplifies the need for multilingual datasets for Natural
Language Processing (NLP) research. However, limited awareness of existing task-specific resources in low-resource languages hinders research. This challenge is especially acute in linguistically diverse countries, such as India. Cross-task profiling of existing Indian speech datasets can alleviate the data scarcity challenge. This involves investigating the utility of datasets across multiple downstream tasks rather than focusing on a single task. Prior surveys typically catalogue datasets for a single task, leaving comprehensive cross-task profiling as an open opportunity. Therefore, we propose Task-Lens, a cross-task survey that assesses the readiness of 50 Indian speech datasets spanning 26 languages for nine downstream speech tasks. First, we analyze which datasets contain metadata and properties suitable for specific tasks. Next, we propose task-aligned enhancements to unlock datasets to their full downstream potential. Finally, we identify tasks and Indian languages that are critically underserved by current resources. Our findings reveal that many Indian speech datasets contain untapped metadata that can support multiple downstream tasks. By uncovering cross-task linkages and gaps, Task-Lens enables researchers to explore the broader applicability of existing datasets and to prioritize dataset creation for underserved tasks and languages.
 \\ \newline  }

\begin{document}

\maketitleabstract

\section{Introduction}
The growing demand for inclusive speech technologies intensifies the need for multilingual datasets for Natural Language Processing (NLP) research. However, most speech datasets are English-centric, so the lack of speech datasets in low-resource languages hinders NLP research. Multilingual speech models aim for inclusivity but often underperform and exhibit linguistic biases in various tasks \citep{jin2020lrspeech}. Consequently, there is an urgent need for task-specific datasets for low-resource languages. This data scarcity challenge can be alleviated by efficient utilization of existing resources. However, researchers are often unaware of existing resources for underrepresented languages, which hinders NLP research for those languages \citep{larasati2025inclusive}. Cross-task profiling is a viable solution to address this problem. Cross-task profiling involves the systematic analysis of dataset attributes to assess their readiness for multiple downstream tasks beyond their originally intended use. 

Cross-task profiling can reveal how existing datasets support diverse tasks. However, most prior works describe dataset creation or catalogue Indian speech resources without profiling them across tasks \citep{ijca2012_indian_speech_review,oaji2016_speech_corpus_indian_asr,ijera2017_speech_corpora_indian,hindi2013_speech_corpora_review,ijert2018_speaker_rec_db,asroil2020}. Recent South Asian surveys present NLP progress (data, models, tasks) across languages with only incidental coverage of speech, and primarily catalogue speech resources without any cross-task profiling \citep{bhaasha2025_southasia_nlp_survey}. As a result, the community often lacks clarity on whether existing speech resources are usable across tasks, especially for underrepresented languages. Although these challenges are global, they are particularly acute in linguistically diverse settings such as India. Cross-task profiling of Indian speech datasets can help researchers maximise the value of available resources and motivate data collection efforts for low-resource languages, which is essential for advancing multilingual speech research.

In this work, we present \textbf{Task-Lens}, a cross-task survey of 50 Indian speech datasets. Beyond cataloging datasets, Task-Lens brings to light resources with utility across multiple tasks that typical AI tools or web searches cannot reliably surface. It highlights Indian speech resources, which include task-aligned metadata that can enable cross-task reuse. Thus, providing readiness profiles that help researchers quickly discover corpora for specific tasks. It also identifies actionable gaps, including missing features that limit cross-task readiness, and pinpoints critically underserved languages and tasks to motivate targeted dataset creation. By reducing discovery time and clarifying reuse pathways, Task-Lens turns scattered resources into a navigable map for advancing multilingual speech research in India. 

We summarize our main contributions below:
\begin{enumerate}[noitemsep]
    \item We present \textbf{Task-Lens}, a cross-task survey that profiles the readiness of Indian speech datasets for multiple downstream tasks using available metadata.
    
    \item Task-Lens includes comprehensive cross-task profiling across nine downstream tasks, using 50 Indian speech datasets covering 26 languages and comprising over 91,257 hours of audio.

    \item We investigate the following research questions: (a) Which speech tasks does each dataset currently support? (b) What improvements would enhance a dataset’s suitability for additional tasks? (c) Which areas of speech research lack adequate dataset support? (d) Which Indian languages offer adequate coverage per task, and where do significant resource gaps remain?
\end{enumerate}

\section{Related Works}

\textbf{Cross-Task Utility Exploration:} Previous works explore profiling frameworks that measure representations across global languages and tasks \citep{yang2021superb, conneau2022fleurs, gebru2021datasheets, chen2024apsipa, mazumder2023dataperf}. However, most surveys and benchmarks have focused on cataloging available datasets \citep{bhaasha2025_southasia_nlp_survey, shilin2018iot, bakhturina2021a}. Thus, several key research questions are underexplored in the literature, such as: (1) While resource papers often associate datasets with specific tasks, rich metadata makes them useful for broader applications. How can we investigate the cross-task utility of existing multilingual speech resources? (2) Is there any urgent need for speech datasets of some specific language for specific tasks? With over 7,000 languages spoken worldwide, addressing these questions can uncover key NLP research gaps. This is especially important for linguistically diverse countries such as India.
This study contributes a cross-task resource analysis that evaluates the readiness of Indian speech datasets for diverse downstream tasks using their documented metadata and properties.

\begin{figure*}[!bht]
  \centering
\includegraphics[scale=0.161]{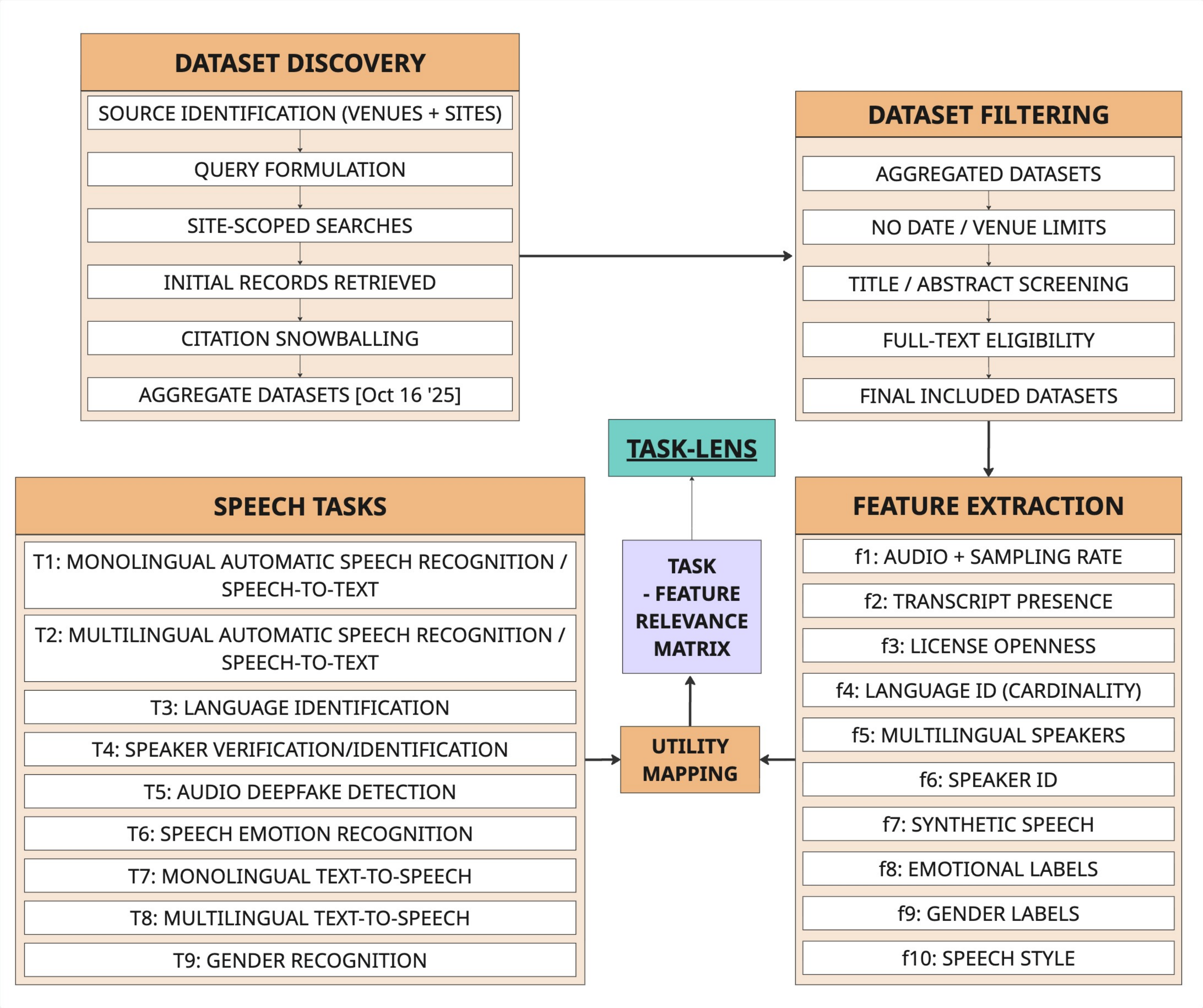}
  \caption{Task-Lens: It involves dataset discovery, dataset filtering, feature extraction, followed by utility mapping that aligns dataset features with task needs via a Task-feature relevance matrix labeled as Required and Optional or Not Applicable. A dataset is `Task-Ready' for a task if it satisfies all 'Required' features for a task. Supported tasks include Automatic Speech Recognition (Monolingual) [T1], Automatic Speech Recognition (Multilingual) [T2], Language Identification [T3], Speaker Verification/Identification [T4], Audio Deepfake Detection [T5], Speech Emotion Recognition [T6], Text-to-Speech (Monolingual) [T7], Text-to-Speech (Multilingual) [T8] and Gender Recognition [T9].}
  \vspace{-1em}
  \label{fig:methodology-flowchart}
\end{figure*}

\textbf{Indian Dataset Landscape:} India’s 22 official languages remain underrepresented in mainstream speech benchmarks, and high-quality, general-purpose corpora are scarce \citep{joshi2020state}. Early surveys cataloged Indian speech resources and noted missing standardization, scale, and metadata consistency \citep{ijca2012_indian_speech_review}. Task-focused reviews address specific areas for Automatic Speech Recognition, Language Identification, code-switching, Text-to-Speech, and Speech Emotion Recognition \citep{asroil2020,unnibhavi2016asr,aarti2018lidreview,dey2022indianlid,sitaram2019codeswitchsurvey,mustafa2022csasrsurvey,agro2025csasrslr,panda2020ttsin,monisha2022ser}. While these efforts advance the coverage of techniques and challenges, the cross-task profiling of existing speech datasets continues to represent an open research opportunity. Furthermore, the closest prior studies to ours are \citet{ijca2012_indian_speech_review} and \citet{bhaasha2025_southasia_nlp_survey}. \citet{ijca2012_indian_speech_review} catalogs Indian speech resources but provides no guidance on repurposing metadata beyond their original intent. Moreover, it includes no resources published after 2012. By contrast, \citet{bhaasha2025_southasia_nlp_survey} presents a timely, well-curated survey of South Asian NLP since 2020, with emphasis on data, models, and tasks. However, this paper is not Indian-speech focused, treats speech only incidentally, and does not articulate speech-specific task-readiness criteria or cross-task reuse pathways. In this work, we move beyond dataset cataloging to perform cross-task profiling of 50 Indian speech datasets covering 26 low-resource Indian languages over nine downstream tasks. Our goal is to highlight available datasets that include sufficient metadata to support tasks beyond their originally intended purpose. This cross-task perspective enables researchers working on underserved languages and speech processing tasks to identify datasets with potential utility for their research. Furthermore, our analysis reveals significant gaps across specific languages and tasks, thereby motivating future data collection efforts targeting these underrepresented areas.

\section{Task-Lens}
\label{sec:tasklensmethodology} 
\textbf{Task-Lens} is a cross-task, utility oriented lens for profiling Indian language speech datasets. The pipeline comprises four stages: dataset discovery, dataset filtering, feature extraction, and utility mapping, as shown in Figure~\ref{fig:methodology-flowchart}. This section reports the methodology in accordance with Preferred Reporting Items for Systematic Reviews and Meta-Analyses (PRISMA) for transparent and reproducible evidence synthesis \citep{Page2021PRISMAEandE}. In line with PRISMA, we detail the  eligibility criteria, information sources, search strategy, selection process, data items, etc.

\textbf{Dataset Discovery:} We searched peer-reviewed venues and dataset registries relevant to speech and language resources in India. Sources included IEEE Xplore, ACM Digital Library, ACL Anthology, ISCA Interspeech, LREC ELRA, Scopus, arXiv, etc.; portals included OpenSLR, Bhashini ULCA and Vatika, AIKosh, LDC IL, ELRA Catalogue, Mozilla Common Voice, Google FLEURS, AI4Bharat repositories, Hugging Face Datasets, Zenodo, and GitHub releases. We used queries that combine resource, task, and language terms (e.g., Indian/Indic/Hindi/Tamil; ``speech dataset''/``speech corpus''; and ``automatic speech recognition''/``language identification''/``text-to-speech''/``speech emotion recognition''/ ``gender recognition''/``audio deepfake detection'') and issued site-scoped searches for each portal. We imposed no date limits or venue filters. We also applied backward and forward citation snowballing from seed papers and dataset pages. The dataset search has been ongoing for several months, with the last comprehensive search conducted on October 16, 2025.

\textbf{Dataset Filtering:} We applied predefined eligibility criteria in two stages: title and abstract screening, followed by full text or webpage review. The first stage removed clear textual records, non-Indic resources, and duplicates. The second stage confirmed the presence of basic metadata, including audio files, sampling rate, and practical accessibility. Inclusion required Indian languages, public documentation with extractable features, and identifiable splits or benchmark usage. Synthetic speech was conditionally included when provenance and generation were documented. 

Following this procedure, we curated a list of 50 Indian speech datasets. The complete list of datasets, along with their abbreviations and references, is presented in Table~\ref{tab:dataset_list}. These datasets cover speech in 26 languages spoken across the Indian demographic population, collectively providing over 91,257 hours of audio. The collection comprises monolingual and multilingual datasets, covering a diverse range of speech styles, domains, speaker types, and annotation depths. 

\begin{table}[!ht]
\centering
\footnotesize
\begin{tabularx}{\linewidth}{p{0.18cm} X}
\toprule
\textbf{ID} & \textbf{Dataset} \\
\midrule
$D_1$ & AccentDB \citep{accentdb} \\
$D_2$ & Assamese TTS Corpus \citep{tamim2025assamese} \\
$D_3$ & BhasaAnuvaad \citep{bhashanuvaad} \\
$D_4$ & Open-source TTS Voices \citep{banga2019indian} \\
$D_5$ & Bengali Numbers Corpus \citep{nahid2018bengali} \\
$D_6$ & South Asian Crowdsourced Speech \\ 
& \citep{kjartansson-etal-sltu2018} \\
$D_7$ & Low-Income Workers \citep{abraham2020crowdsourcing} \\
$D_8$ & FLEURS \citep{conneau2022fleurs} \\
$D_9$ & GACMIS Songs \citep{ujjwalll2020gacmis} \\
$D_{10}$ & GlobalPhone Speaker Package \citep{elra-s0400-2018} \\
$D_{11}$ & GRAM VAANI \citep{gramvaani2022} \\
$D_{12}$ & Hindi-Tamil-English ASR \citep{speechlab_iitm_2021_asr} \\
$D_{13}$ & Indian Folk Music \citep{singh2022indian} \\
$D_{14}$ & Regional Music \citep{singh2021indianregional} \\
$D_{15}$ & Indic TTS (IITM) \citep{iitm2025indictts} \\
$D_{16}$ & IndicSpeech (TTS) \citep{iitm_indicspeech} \\
$D_{17}$ & IndicSUPERB \citep{javed2023indicsuperb} \\
$D_{18}$ & IndicVoices-R \citep{indicvoices_r} \\
$D_{19}$ & Kashmiri Data Corpus \citep{openslr122} \\
$D_{20}$ & KritiSamhita \citep{kritisamhita2020} \\
$D_{21}$ & Lahaja \citep{lahaja2024} \\
$D_{22}$ & MS Indic Speech \citep{msindicspeech2024} \\
$D_{23}$ & MUCS Site \citep{diwan2021mucs} \\
$D_{24}$ & Nexdata AI 759h \citep{nexdata2025hindi} \\
$D_{25}$ & NISP \citep{nisp2021} \\
$D_{26}$ & NPTEL2020 \citep{nptel2020} \\
$D_{27}$ & Opensource Multispeaker Data \citep{gujarati_telugu_tts} \\
$D_{28}$ & Rajasthani Hindi (MS) \citep{msrajasthani2024} \\
$D_{29}$ & Rasa \citep{rasa_dataset} \\
$D_{30}$ & SMC Malayalam \citep{smc_malayalam} \\
$D_{31}$ & Svarah \citep{javed2023svarah} \\
$D_{32}$ & Urdu Recognition (Desktop) \citep{ELRA2024UrduDesktop} \\
$D_{33}$ & Vāksañcayaḥ \citep{vaksanca} \\
$D_{34}$ & IndicSynth \citep{indicsynth} \\
$D_{35}$ & NIRANTAR \citep{nirantar2025interspeech} \\
$D_{36}$ & Shrutilipi-Anuvaad \citep{shrutilipi2025anuvaad} \\
$D_{37}$ & IIITH-HE-CM \citep{rambabu2018iiithhecm} \\
$D_{38}$ & EmoTa: A Tamil Emotional Speech Dataset \citep{thevakumar2025emota} \\
$D_{39}$ & IndicFake \citep{anonymous2025indicfake} \\
$D_{40}$ & SPIRE-SIES \citep{prabhu2023spiresies} \\
$D_{41}$ & BanglaSER \citep{das2022bangalser} \\
$D_{42}$ & SUBESCO \citep{sultana2021subesco} \\
$D_{43}$ & KBES \citep{billah2023kbes} \\
$D_{44}$ & BanSpEmo \citep{sultana2025banspemo} \\
$D_{45}$ & SEA\_Spoof \citep{wu2025seaspoofbridginggapmultilingual} \\
$D_{46}$ & IIITH-ILSC \citep{vuddagiri2018iiith-ilsc} \\
$D_{47}$ & I-MSV \citep{mishra2023imsv2022indicmultilingualmultisensor} \\
$D_{48}$ & IndieFake \citep{kumar2025indiefakedatasetbenchmarkdataset} \\
$D_{49}$ & Bangla Speech Corpus \citep{ahmed2020banglacorpus} \\
$D_{50}$ & Indic-TEDST \citep{sethiya2024indictedst} \\
\bottomrule
\end{tabularx}
\vspace{-1em}
\caption{List of the 50 Indian speech datasets included in our analysis. Each dataset is assigned a unique identifier ($D_1$–$D_{50}$), which is used consistently throughout the paper. The table includes established and lesser-known resources, along with relevant citations.}
\label{tab:dataset_list}
\end{table}

\textbf{Feature Extraction:} After filtering, we extracted 10 descriptive features for each dataset using a standardized schema. Table~\ref{features_rationale} summarizes the selected features and their rationale. Each feature will be referenced as presented in the table, in the form \(f_i\)\footnote{All per dataset feature values will be released upon publication.}.

\textbf{Utility Mapping: }
After the feature extraction stage, we get a set of features for each dataset. The next step is to perform a task utility mapping for these datasets. Each task will be referenced in the form \(T_j\). For utility mapping, we considered nine representative speech-technology tasks:
\[
\begin{aligned}
T_1 &:\ \text{Monolingual Automatic Speech} \\
 & \hspace{0.1in}\ \text{Recognition / Speech-to-Text} \\
  & \hspace{0.1in}\ \text{(MO-ASR/STT)} \\
T_2 &:\ \text{Multilingual Automatic Speech} \\
 & \hspace{0.1in}\ \text{Recognition / Speech-to-Text} \\
  & \hspace{0.1in}\ \text{(ML-ASR/STT)} \\
T_3 &:\ \text{Language Identification (LID)} \\
T_4 &:\ \text{Speaker Verification/Identification (SV/SID)} \\
T_5 &:\ \text{Audio Deepfake Detection (ADD)} \\
T_6 &:\ \text{Speech Emotion Recognition (SER)} \\
T_7 &:\ \text{Monolingual Text-to-Speech (MO-TTS)} \\
T_8 &:\ \text{Multilingual Text-to-Speech (ML-TTS)} \\
T_9 &:\ \text{Gender Recognition (GRE)} \\
\end{aligned}
\]

Here, $T_1$ - $T_2$ and $T_7$ - $T_8$ are treated separately to reflect the difference between generating text/speech across multiple languages and producing high-quality output in a single language. 

\textbf{Task–Feature Mapping:}
The task–feature relevance matrix in Table~\ref{tab:feature_task_requirements} reflects a rule-based mapping grounded in task literature and benchmark specifications. For each pair \((f_i, T_j)\), we first formalized the task and its core data needs. For example, LID prioritizes utterance level language labels rather than transcripts. We then validated requirements using benchmark documentation and recent surveys across Indian speech research \citep{panda2020ttsin,monisha2022ser, oaji2016_speech_corpus_indian_asr, aarti2018lidreview, asroil2020, bhaasha2025_southasia_nlp_survey, dey2022indianlid}. We minimized required features to ensure consistent and fair utility judgments across datasets and tasks. The mapping remains extensible as task definitions evolve or new task families emerge.

For each task \(t\) and feature \(f\), we assign a categorical label \(\{\cmark,\qmark\}\), where \cmark\ denotes Required, \qmark\ denotes Optional or Not Required. Table~\ref{tab:feature_task_requirements} summarizes these labels for all combinations of tasks and features.
\begin{table}[H]
\centering
\footnotesize
\setlength{\tabcolsep}{0 pt}
\begin{tabular}{l r}
\hline
\textbf{Feature}  & \textbf{Rationale} \\ 
\hline
$f_1$: Audio +  & Ensures audio fidelity  \\
\hspace{0.17in} Sampling Rate & and robustness \\
$f_2$: Transcript Presence & Enables text-speech alignment \\
$f_3$: License Openness & Governs accessibility reuse \\
$f_4$: Language ID & Supports multilingual tasks \\
\hspace{0.17in} (Cardinality)   &  \\
$f_5$: Multilingual & Facilitates code-switching \\
\hspace{0.17in} Speakers & research \\
$f_6$: Speaker ID & Essential for speaker-centric  \\
\hspace{0.17in} & tasks \\
$f_7$: Synthetic Speech & Used for augmentation/detection  \\
$f_8$: Emotional Labels & Supports affective computing \\
$f_9$: Gender Labels & Enables fairness/robustness\\
& analyses  \\
$f_{10}$: Speech Style & Distinguishes read vs. \\
& conversational/scripted,  \\
& affecting task transferability \\

\hline
\end{tabular}
\caption{Utility feature summary.}
\label{features_rationale}
\end{table}

\section{Task-Lens Utility Exploration}
\subsection{Cross-Task Dataset Utility}
Standard practice in speech processing involves designing each dataset ~$d$ for a single task ~$t$. However, published datasets often contain rich metadata, making them suitable for other tasks too. For instance, Librispeech was introduced for the ASR task but has been leveraged for speaker verification task \citep{panayotov2015librispeech, sharma-buduru-2022-fatnet}. The cross-task utility of published datasets is an area that remains underexplored. Consequently, due to a lack of cross-task utility exploration, published datasets are often underutilized for diverse applications despite their potential utility. 
\begin{table*}[!hbt]
\centering
\footnotesize
\begin{tabular}{p{3.3cm}*{9}{c}}
\hline
\textbf{Feature ($f_i$)} & \textbf{MO-ASR/} & \textbf{ML-ASR/} & \textbf{LID} & \textbf{SV/SID} & \textbf{ADD} & \textbf{SER} & \textbf{MO-TTS} & \textbf{ML-TTS} & \textbf{GRE}\\
 & \textbf{STT} & \textbf{STT} &  &  &  &  &  &  & \\
\hline
$f_1$: Audio +  & \cmark & \cmark & \cmark & \cmark & \cmark & \cmark & \cmark & \cmark & \cmark \\
\hspace{0.17in} Sampling Rate  &  &  &  &  &  &  &  &  & \\
$f_2$: Transcript Presence      & \cmark & \cmark & \qmark & \qmark & \qmark & \qmark & \cmark & \cmark & \qmark \\
$f_3$: License Openness        & \qmark & \qmark & \qmark & \qmark & \qmark & \qmark & \qmark & \qmark & \qmark \\
$f_4$: Language ID             & \qmark & \cmark & \cmark & \qmark & \qmark & \qmark & \qmark & \cmark & \qmark \\
\hspace{0.17in} (Cardinality)   &  &  &  &  &  &  &  &  & \\
$f_5$: Multilingual   & \qmark & \cmark & \qmark & \qmark & \qmark & \qmark & \qmark & \cmark & \qmark \\
\hspace{0.17in} Speakers   &  &  &  &  &  &  &  &  & \\
$f_6$: Speaker ID              & \qmark & \qmark & \qmark & \cmark & \qmark & \qmark & \qmark & \qmark & \qmark \\
$f_7$: Synthetic Speech        & \qmark & \qmark & \qmark & \qmark & \cmark & \qmark & \qmark & \qmark & \qmark \\
$f_8$: Emotional Labels          & \qmark & \qmark & \qmark & \qmark & \qmark & \cmark & \qmark & \qmark & \qmark \\
$f_9$: Gender Labels        & \qmark & \qmark & \qmark & \qmark & \qmark & \qmark & \qmark & \qmark & \cmark \\
$f_{10}$: Speech Style        & \qmark & \qmark & \qmark & \qmark & \qmark & \qmark & \qmark & \qmark & \qmark \\
\hline
\end{tabular}
\caption{Task–feature relevance matrix for dataset–task screening. Symbols: \cmark\ = Required, \qmark\ = Optional / Not Required. For each task, entries marked \cmark\ form the minimal, exhaustive set of required features under standard task definitions.}
\label{tab:feature_task_requirements}
\end{table*}

The lack of cross-task studies creates additional challenges for NLP researchers working on low-resource language tasks. Therefore, we use Task-Lens to address the following research questions: \textit{(1) Which tasks \(t\) does each dataset \(d\) currently support? (2) What enhancements would make a dataset suitable for cross-task applications?}

\textbf{Setup.} For each dataset \(d\) and task \(t\), Task-Lens checks whether all features marked \cmark\ (Required) in the task-feature relevance matrix (Table~\ref{tab:feature_task_requirements}) are present. A dataset is `Task-Ready' for a task if it satisfies all `Required' features for a task. Task and dataset details appear in Section~\ref{sec:tasklensmethodology} and Table~\ref{tab:dataset_list}.


\begin{table*}[!htb]
\centering
\footnotesize
\begin{tabular}{lrrrrrrrrr}
\toprule
\textbf{Dataset} & \textbf{$ T_1 $} & \textbf{$ T_2 $} & \textbf{$ T_3 $} & \textbf{$ T_4 $} & \textbf{$ T_5 $} & \textbf{$ T_6 $} & \textbf{$ T_7 $} & \textbf{$ T_8 $} & \textbf{$ T_9 $} \\
\hline
$D_{1}$ & $f_2$ & $f_2$, $f_{5}$ & \checkmark & \checkmark & $f_7$ & $f_8$ & $f_2$ & $f_2$, $f_{5}$ & \checkmark \\
$D_{2}$ & \checkmark & $f_5$ & \checkmark & $f_6$ & $f_7$ & $f_8$ & \checkmark & $f_5$ & \checkmark \\
$D_{3}$ & \checkmark & \checkmark & \checkmark & $f_6$ & $f_7$ & $f_8$ & \checkmark & \checkmark & \checkmark \\
$D_{4}$ & \checkmark & \checkmark & \checkmark & \checkmark & $f_7$ & $f_8$ & \checkmark & \checkmark & \checkmark \\
$D_{5}$ & \checkmark & $f_5$ & \checkmark & $f_6$ & $f_7$ & $f_8$ & \checkmark & $f_5$ & $f_9$ \\
$D_{6}$ & \checkmark & \checkmark & \checkmark & \checkmark & $f_7$ & $f_8$ & \checkmark & \checkmark & \checkmark \\
$D_{7}$ & \checkmark & $f_5$ & \checkmark & $f_6$ & $f_7$ & $f_8$ & \checkmark & $f_5$ & \checkmark \\
$D_{8}$ & \checkmark & \checkmark & \checkmark & $f_6$ & $f_7$ & $f_8$ & \checkmark & \checkmark & \checkmark \\
$D_{9}$ & \checkmark & \checkmark & \checkmark & $f_6$ & $f_7$ & $f_8$ & \checkmark & \checkmark & \checkmark \\
$D_{10}$ & \checkmark & \checkmark & \checkmark & $f_6$ & $f_7$ & $f_8$ & \checkmark & \checkmark & $f_9$ \\
$D_{11}$ & \checkmark & $f_5$ & \checkmark & $f_6$ & $f_7$ & $f_8$ & \checkmark & $f_5$ & \checkmark \\
$D_{15}$ & \checkmark & \checkmark & \checkmark & \checkmark & $f_7$ & $f_8$ & \checkmark & \checkmark & \checkmark \\
$D_{16}$ & \checkmark & \checkmark & \checkmark & \checkmark & $f_7$ & $f_8$ & \checkmark & \checkmark & \checkmark \\
$D_{17}$ & \checkmark & \checkmark & \checkmark & $f_6$ & $f_7$ & $f_8$ & \checkmark & \checkmark & \checkmark \\
$D_{18}$ & \checkmark & \checkmark & \checkmark & \checkmark & $f_7$ & $f_8$ & \checkmark & \checkmark & \checkmark \\
$D_{19}$ & \checkmark & $f_5$ & \checkmark & \checkmark & $f_7$ & $f_8$ & \checkmark & $f_5$ & $f_9$ \\
$D_{21}$ & \checkmark & $f_5$ & \checkmark & $f_6$ & $f_7$ & $f_8$ & \checkmark & $f_5$ & \checkmark \\
$D_{22}$ & \checkmark & \checkmark & \checkmark & \checkmark & $f_7$ & $f_8$ & \checkmark & \checkmark & \checkmark \\
$D_{23}$ & \checkmark & \checkmark & \checkmark & $f_6$ & $f_7$ & $f_8$ & \checkmark & \checkmark & \checkmark \\
$D_{24}$ & \checkmark & $f_5$ & \checkmark & \checkmark & $f_7$ & $f_8$ & \checkmark & $f_5$ & \checkmark \\
$D_{25}$ & \checkmark & \checkmark & \checkmark & $f_6$ & $f_7$ & $f_8$ & \checkmark & \checkmark & \checkmark \\
$D_{26}$ & \checkmark & $f_5$ & \checkmark & $f_6$ & $f_7$ & $f_8$ & \checkmark & $f_5$ & \checkmark \\
$D_{27}$ & \checkmark & \checkmark & \checkmark & $f_6$ & $f_7$ & $f_8$ & \checkmark & \checkmark & \checkmark \\
$D_{28}$ & \checkmark & $f_5$ & \checkmark & \checkmark & $f_7$ & $f_8$ & \checkmark & $f_5$ & \checkmark \\
$D_{29}$ & \checkmark & \checkmark & \checkmark & $f_6$ & $f_7$ & \checkmark & \checkmark & \checkmark & \checkmark \\
$D_{30}$ & \checkmark & $f_5$ & \checkmark & \checkmark & $f_7$ & $f_8$ & \checkmark & $f_5$ & \checkmark \\
$D_{31}$ & \checkmark & $f_5$ & \checkmark & $f_6$ & $f_7$ & $f_8$ & \checkmark & $f_5$ & \checkmark \\
$D_{32}$ & $f_2$ & $f_2$, $f_5$ & \checkmark & $f_6$ & $f_7$ & $f_8$ & $f_2$ & $f_2$, $f_5$ & \checkmark \\
$D_{33}$ & \checkmark & $f_5$ & \checkmark & $f_6$ & $f_7$ & $f_8$ & \checkmark & $f_5$ & \checkmark \\
$D_{34}$ & \checkmark & \checkmark & \checkmark & - & \checkmark & $f_8$ & \checkmark & \checkmark & \checkmark \\
$D_{35}$ & \checkmark & \checkmark & \checkmark & \checkmark & $f_7$ & $f_8$ & \checkmark & \checkmark & \checkmark \\
$D_{36}$ & \checkmark & \checkmark & \checkmark & $f_6$ & $f_7$ & $f_8$ & \checkmark & \checkmark & \checkmark \\
$D_{37}$ & $f_2$ & $f_2$ & \checkmark & \checkmark & $f_7$ & $f_8$ & $f_2$ & $f_2$ & \checkmark \\
$D_{38}$ & \checkmark & $f_5$ & \checkmark & \checkmark & $f_7$ & \checkmark & \checkmark & $f_5$ & \checkmark \\
$D_{39}$ & $f_2$ & $f_2$ & \checkmark & $f_6$ & \checkmark & $f_8$ & $f_2$ & $f_2$ & \checkmark \\
$D_{40}$ & \checkmark & $f_5$ & \checkmark & \checkmark & $f_7$ & $f_8$ & \checkmark & $f_5$ & \checkmark \\
$D_{41}$ & \checkmark & $f_5$ & \checkmark & \checkmark & $f_7$ & \checkmark & \checkmark & $f_5$ & \checkmark \\
$D_{42}$ & \checkmark & $f_5$ & \checkmark & \checkmark & $f_7$ & \checkmark& \checkmark & $f_5$ & \checkmark \\
$D_{43}$ & $f_2$ & $f_2$, $f_5$ & \checkmark & \checkmark & $f_7$ & \checkmark & $f_2$ & $f_2$, $f_5$ & \checkmark \\
$D_{44}$ & \checkmark & $f_5$ & \checkmark & $f_6$ & $f_7$ & \checkmark & \checkmark & $f_5$ & \checkmark \\
$D_{45}$ & $f_2$ & $f_2$ & \checkmark & $f_6$ & \checkmark & $f_8$ & $f_2$ & $f_2$ & \checkmark \\
$D_{46}$ & $f_2$ & $f_2$ & \checkmark & \checkmark & $f_7$ & $f_8$ & $f_2$ & $f_2$ & \checkmark \\
$D_{47}$ & $f_2$ & $f_2$ & \checkmark &  \checkmark & $f_7$ & $f_8$ & $f_2$ & $f_2$ & $f_9$ \\
$D_{48}$ & \checkmark & $f_5$ & \checkmark & - & \checkmark & $f_8$ & \checkmark & $f_5$ & \checkmark \\
$D_{49}$ & \checkmark & $f_5$ & \checkmark & - & \checkmark & $f_8$ & \checkmark & $f_5$ & \checkmark \\
$D_{50}$ & \checkmark & \checkmark & \checkmark & \checkmark & $f_7$ & $f_8$ & \checkmark & \checkmark & $f_9$ \\
\bottomrule
\end{tabular}
\caption{Dataset–task readiness summary. A \checkmark\ denotes Task-Ready (all required features present); missing required features are listed as \(f_i,f_j\).
Tasks: \(T_1\)=MO-ASR/STT, \(T_2\)=ML-ASR/STT, \(T_3\)=LID, \(T_4\)=SV/SID, \(T_5\)=ADD, \(T_6\)=SER, \(T_7\)=MO-TTS, \(T_8\)=ML-TTS, \(T_9\)=GRE. 
\emph{Notes:} Datasets \(D_{13}\), \(D_{14}\), and \(D_{20}\) are excluded because they are music-oriented rather than speech. Dataset \(D_{12}\) is a challenge release with extensive hours but insufficient public metadata for profiling, so it is omitted here. Although dataset \(D_{34}\), \(D_{48}\), and \(D_{49}\) includes \(f_6\), which helps map a real speaker's voice to its synthetic counterpart, the dataset is better suited for anti-spoofing tasks than for speaker verification, where real and synthetic voices are matched.}
\label{tab:table_utility}
\end{table*}

\textbf{Observations:} As shown in Table~\ref{tab:table_utility}, the corpus set demonstrates broad task coverage across datasets. We observed that datasets $D_{4}$, $D_{6}$, $D_{15}$, $D_{16}$, $D_{18}$, $D_{22}$, $D_{29}$, $D_{34}$, and $D_{35}$ contain the required features to support seven of the nine tasks. These datasets commonly lack speaker identifiers ($f_6$), synthetic speech ($f_7$), or emotion labels ($f_8$), which are essential for extending usability to speaker verification ($T_4$), deepfake detection ($T_5$), and emotion recognition ($T_6$). All of them would reach complete coverage by incorporation of the above. Furthermore, multiple \textit{Task-Ready} datasets qualify for specific objectives and fall short of only a few features in supporting additional tasks. From a development perspective, several datasets represent clear candidates for improvement, where targeted inclusion of these key metadata would substantially enhance their cross‑task readiness.

\subsection{Task-Wise Data Requirement}
Developing robust speech models for Indian languages requires a clear understanding of the existing datasets for each task and identifying where gaps persist. Although ASR, TTS, and GRE resources have expanded rapidly (Table~\ref{tab:table_utility}), the distribution of datasets across tasks remains uneven and opaque to practitioners  \citep{javed2023indicsuperb,indicvoices_r}. Researchers often spend significant effort surveying repositories to determine whether tasks such as emotion recognition or deepfake detection have sufficient data, but find limited or no Indian-specific resources  \citep{busso2008iemocap}. These observations motivate our second inquiry: \textit{Which tasks lack sufficient dataset support for the Indian population?}

\begin{figure}[!htbp]
  \centering
  \includegraphics[scale=0.2]{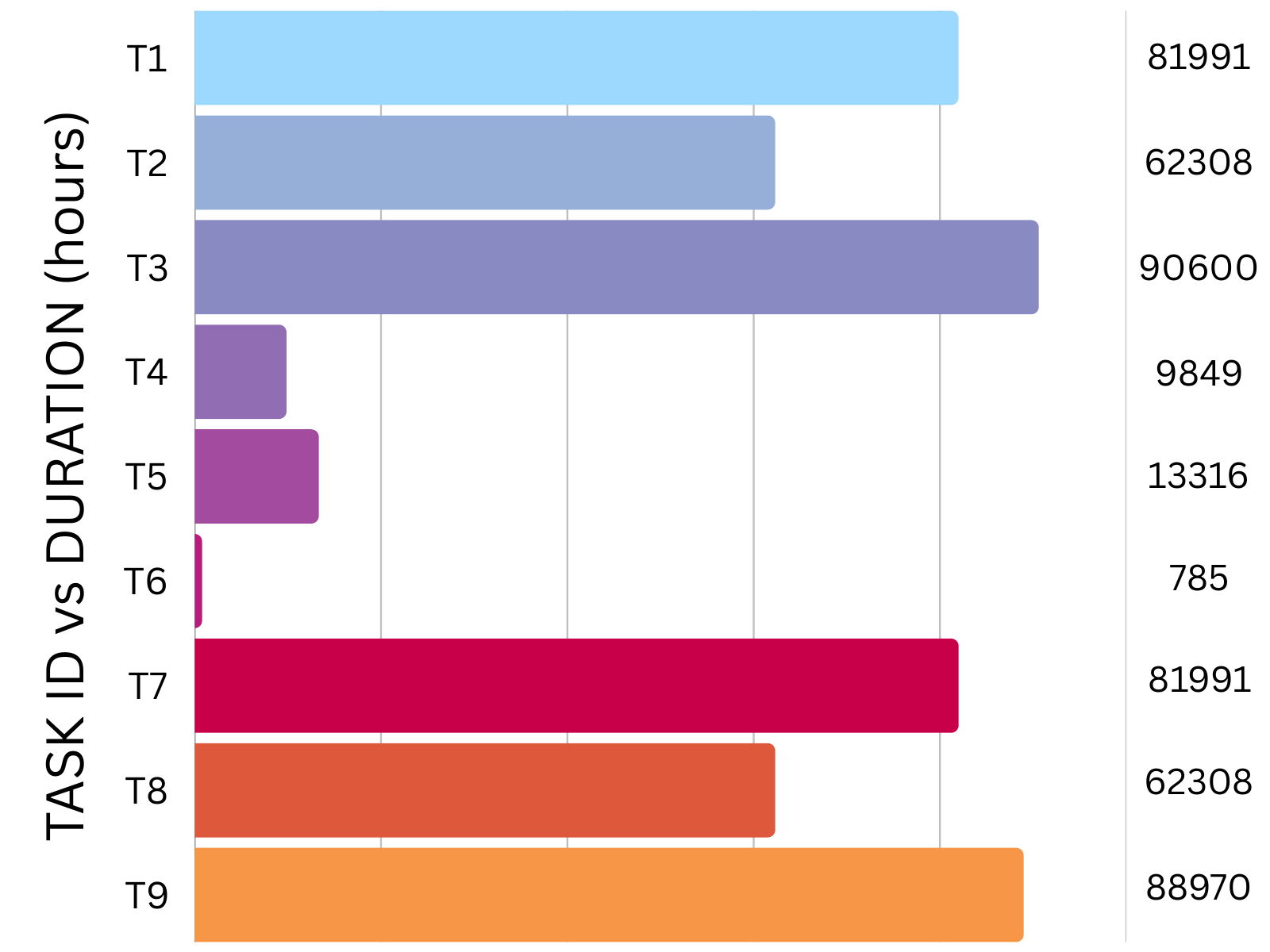}
  \caption{Distribution of total dataset duration for each task in hours for direct comparison. There is an urgent need of datasets for tasks $T_4$ (SV/SID), $T_5$ (ADD), and $T_6$ (SER).}
  \label{fig:task-wise datasets}
\end{figure}

\textbf{Setup:} We used Task-Lens outputs (Table~\ref{tab:table_utility}) to select tasks for each dataset. Next, we compiled, for each task, a list of datasets that satisfy it, yielding multiple Task-Ready datasets per task. After identifying these datasets, we examined their total speech duration (Figure~\ref{fig:task-wise datasets}) to assess coverage.

\textbf{Observations:} Figure~\ref{fig:task-wise datasets} shows that tasks $T_3$ (LID) and $T_9$ (GRE) achieve the highest coverage at roughly 90,000 hours, reflecting support from multiple large scale datasets and feature requirements that are common across many corpora. Tasks $T_1$ (MO-ASR/STT), $T_2$ (ML-ASR/STT), $T_7$ (MO-TTS) and $T_8$ (ML-TTS) each reach about 60,000 hours, indicating reasonable, though still improvable, support driven by overlapping feature needs. In contrast, tasks $T_4$ (SV/SID) and $T_5$ (ADD) cover only about 9,000 to 13,000 hours, underscoring the frequent absence of speaker identifiers and synthetic speech in low resource datasets. Finally, there is an urgent need for emotion labels in Indian language speech corpora, as only 785 hours currently support SER. Some examples of datasets with highest duration for $T_1$ and $T_7$ are: $D_3$, $D_{26}$, $D_{34}$, $D_{35}$, $D_{18}$; for $T_2$ and $T_8$: $D_3$, $D_{34}$, $D_{35}$, $D_{18}$, $D_{17}$; for $T_3$ and $T_9$: $D_3$, $D_{26}$, $D_{39}$, $D_{34}$, $D_{35}$; for $T_4$: $D_{35}$, $D_{18}$, $D_{6}$, $D_{22}$, $D_{24}$; for $T_5$: $D_{39}$, $D_{34}$, $D_{49}$, $D_{45}$, $D_{48}$; and for $T_6$ are: $D_{29}$, $D_{41}$, $D_{42}$, $D_{44}$, $D_{38}$.

\subsection{Linguistic Data Requirement}

Researchers in the NLP community strive to develop multilingual and inclusive speech technologies. However, the majority of speech datasets are in English. Consequently, the lack of low-resource language datasets hinders the development of multilingual speech technologies. Only a few of the world's languages (approx 7,000) have sufficient resources for human language technologies \citep{besacier2014automatic}. This gap motivates our third inquiry: \textit{Which Indian languages have adequate dataset support for each task, and where do critical language-specific gaps persist?}

\begin{figure}[!ht]
  \centering
  \includegraphics[scale=0.225]{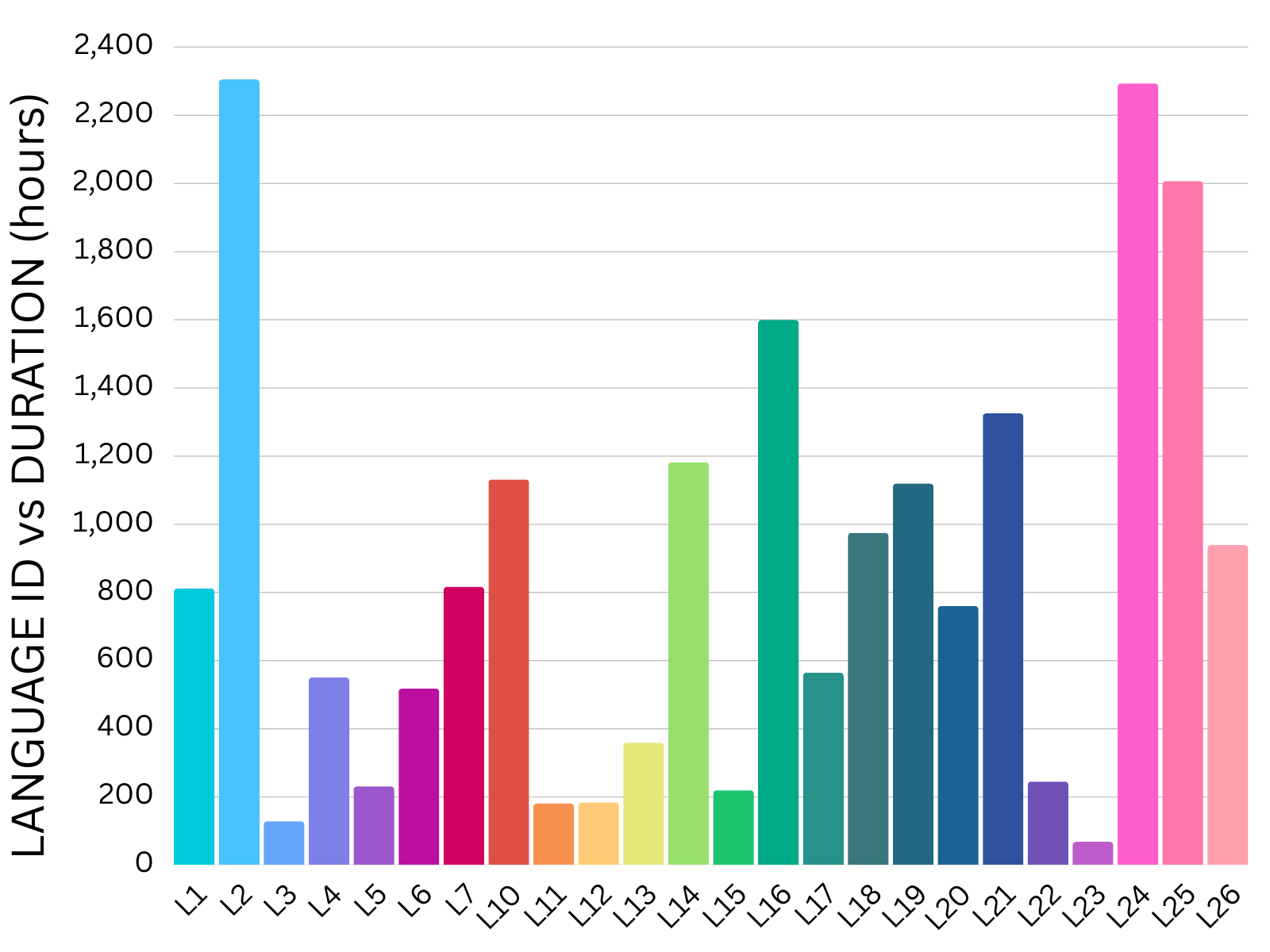}
  \caption{Total speech duration for each Indian language (L$_1$–L$_{26}$) across all 50 datasets. Language L$_8$ (Hindi) and L$_9$ (Indian English) have 3,981 and 16,154 hours of data and were excluded from the figure due to duration; they would have dominated the visualization and obscured relative differences among datasets. Languages like $L_{2}$, $L_{24}$, and $L_{25}$ have the highest duration, whereas languages like $L_{23}$, $L_{3}$, $L_{11}$, and $L_{12}$ are virtually absent.}
  \label{fig:language_duration_line}
\end{figure}

\textbf{Setup:} For each task, we selected all the Task-Ready datasets listed in  Table~\ref{tab:table_utility}. Next, we identified the languages present in those datasets, encompassing a total of \textit{26 languages}. Next, we calculated total audio duration of each language across these datasets. Table~\ref{tab:language_task_durations} shows the total speech durations (in hours) for Task-Ready datasets across 26 languages and nine downstream tasks (\(T_{1}\)–\(T_{9}\)). The notation for each language (L$_1$--L$_{26}$) corresponds to the language identifiers defined in Table~\ref{tab:language_task_durations}, which lists their full names and associated datasets for clarity. Table~\ref{tab:language_task_durations} also lists the IDs of all curated datasets containing recordings in each language, guiding researchers to their relevant data resources. Figure~\ref{fig:language_duration_line} shows the speech duration per language across all 50 datasets.

\begin{table*}[!htb]
\centering
\small
\setlength{\tabcolsep}{2.5pt} 
\begin{tabular}{@{}lrrrrrrrrr p{0.30\textwidth}@{}}
\toprule
\textbf{Language} & \textbf{$T_1$} & \textbf{$T_2$} & \textbf{$T_3$} & \textbf{$T_4$} &
\textbf{$T_5$} & \textbf{$T_6$} & \textbf{$T_7$} & \textbf{$T_8$} & \textbf{$T_9$} & \textbf{Datasets} \\
\midrule
$L_{1}$: Assamese        & 537 & 492 & \cellcolor{midgreen}806 & 465 & 264 & \cellcolor{midred} 17 & 537 & 492 & \cellcolor{midgreen} 806 &
  \begin{tabular}[t]{@{}l@{}} $D_{i\in\{2,3,8,15,18,29,35,39,46\}}$ \\ \end{tabular} \\
$L_{2}$: Bengali         & 1990 & 1015 & \cellcolor{midgreen}\cellcolor{midgreen}2297 & 635 & 1418 & \cellcolor{midred} 28 & 1990 & 1015 & 2280 &
    \begin{tabular}[t]{@{}l@{}} $D_{i\in\{3\!-\!6,8,15\!-\!18,29,34\!-\!36, 39,41\!-\!44\}}$ \\ \end{tabular} \\
         &  &  &  &  & & &  &  & & 
\begin{tabular}[t]{@{}l@{}} $D_{i\in\{49, 50\}}$ \\ \end{tabular} \\
$L_{3}$: Bhojpuri        & \cellcolor{midgreen}120 & \cellcolor{midgreen}120 & \cellcolor{midgreen}120 & \cellcolor{midred} 0 & \cellcolor{midred} 0 & \cellcolor{midred} 0 & \cellcolor{midgreen}120 & \cellcolor{midgreen}120 & \cellcolor{midgreen}120 &
\begin{tabular}[t]{@{}l@{}} $D_{i\in\{9\}}$ \\ \end{tabular} \\
$L_{4}$: Bodo            & 483 & 483 & \cellcolor{midgreen}550 & 487 & 63 & \cellcolor{midred} 0 & 483 & 483 & \cellcolor{midgreen}550 &
\begin{tabular}[t]{@{}l@{}} $D_{i\in\{15,18,35,39,46\}}$ \\ \end{tabular} \\
$L_{5}$: Dogri           & 200 & 200 & \cellcolor{midgreen}231 & 204 & 26 & \cellcolor{midred} 0 & 200 & 200 & \cellcolor{midgreen}231 &
\begin{tabular}[t]{@{}l@{}} $D_{i\in\{15,18,35,39,46\}}$ \\ \end{tabular} \\
$L_{6}$: Garhwali        & 140 & 140 & \cellcolor{midgreen}515 & 25 & 371 & \cellcolor{midred} 0 & 140 & 140 & \cellcolor{midgreen}515 &
\begin{tabular}[t]{@{}l@{}} $D_{i\in\{9,35,39,46\}}$ \\ \end{tabular} \\
$L_{7}$: Gujarati        & \cellcolor{midgreen}810 & \cellcolor{midgreen}810 & \cellcolor{midgreen}810 & 409 & 197 & \cellcolor{midred} 0 & \cellcolor{midgreen}810 & \cellcolor{midgreen}810 & 799 &
\begin{tabular}[t]{@{}l@{}} $D_{i\in\{3,8,15,17,18,22,23,27,34,50\}}$ \\ \end{tabular} \\
$L_{8}$: Hindi           & 3126 & 1243 & 3792 & 1138 & 1001 & \cellcolor{midred} 0 & 3126 & 1243 & \cellcolor{midgreen}3689 &
\begin{tabular}[t]{@{}l@{}} $D_{i\in\{3,8,9,11,15\!-\!18,21,23\!-\!25\}}$ \\ \end{tabular} \\
         &  &  &  &  & & &  &  & &
\begin{tabular}[t]{@{}l@{}} $D_{i\in\{34\!-\!36,39,45,46,50\}}$ \\ \end{tabular} \\
$L_{9}$: Indian English  & 15940 & 32 & \cellcolor{midgreen}15964 & 184 & 27 & \cellcolor{midred} 0 & 15940 & 32 & \cellcolor{midgreen}15964 &
\begin{tabular}[t]{@{}l@{}} $D_{i\in\{1,25,26,31,40,46,48\}}$ \\ \end{tabular} \\
$L_{10}$: Kannada        & 550 & 550 & \cellcolor{midgreen}1124 & 195 & 837 & \cellcolor{midred} 0 & 550 & 550 & 1121 &
\begin{tabular}[t]{@{}l@{}} $D_{i\in\{3,8,15,17,18,25,27,34,35,39,46,50\}}$ \\ \end{tabular} \\
$L_{11}$: Kashmiri       & 174 & 171 & \cellcolor{midgreen}179 & \cellcolor{midgreen}179 & \cellcolor{midred} 0 & \cellcolor{midred} 0 & 174 & 171 & 176 &
\begin{tabular}[t]{@{}l@{}} $D_{i\in\{18,19,35,46\}}$ \\ \end{tabular} \\
$L_{12}$: Konkani        & 177 & 177 & \cellcolor{midgreen}182 & \cellcolor{midgreen}182 & \cellcolor{midred} 0 & \cellcolor{midred} 0 & 177 & 177 & \cellcolor{midgreen}182 &
\begin{tabular}[t]{@{}l@{}} $D_{i\in\{15,18,35,46\}}$ \\ \end{tabular} \\
$L_{13}$: Maithili       & 349 & 349 & \cellcolor{midgreen}359 & 354 & 5 & \cellcolor{midred} 0 & 349 & 349 & \cellcolor{midgreen}359 &
\begin{tabular}[t]{@{}l@{}} $D_{i\in\{15,18,35,39,46\}}$ \\ \end{tabular} \\
$L_{14}$: Malayalam      & 681 & 679 & \cellcolor{midgreen}1180 & 336 & 590 & \cellcolor{midred} 0 & 681 & 679 & 1174 &
\begin{tabular}[t]{@{}l@{}} $D_{i\in\{3,8,15\!-\!18,25,27,30,34\!-\!36,39,46,50\}}$ \\ \end{tabular} \\
$L_{15}$: Manipuri       & 100 & 100 & \cellcolor{midgreen}218 & 104 & 113 & \cellcolor{midred} 0 & 100 & 100 & \cellcolor{midgreen}218 &
\begin{tabular}[t]{@{}l@{}} $D_{i\in\{15,18,35,39,46\}}$ \\ \end{tabular} \\
$L_{16}$: Marathi        & 1103 & 994 & \cellcolor{midgreen}1594 & 216 & 965 & \cellcolor{midred} 0 & 1103 & 994 & 1570 &
\begin{tabular}[t]{@{}l@{}} $D_{i\in\{3,7,8,15,17,18,23,27,34,35,39,46,50\}}$ \\ \end{tabular} \\
$L_{17}$: Nepali         & 558 & 558 & \cellcolor{midgreen}563 & 548 & \cellcolor{midred} 0 & \cellcolor{midred} 0 & 558 & 558 & 563 &
\begin{tabular}[t]{@{}l@{}} $D_{i\in\{3,4,6,8,15,18,35,46\}}$ \\ \end{tabular} \\
$L_{18}$: Odia           & 650 & 650 & \cellcolor{midgreen}973 & 208 & 410 & \cellcolor{midred} 0 & 650 & 650 & \cellcolor{midgreen}973 &
\begin{tabular}[t]{@{}l@{}} $D_{i\in\{3,8,15,17,18,23,34\!-\!36,39,46\}}$ \\ \end{tabular} \\
$L_{19}$: Punjabi        & 765 & 765 & \cellcolor{midgreen}1115 & 228 & 728 & \cellcolor{midred} 0 & 765 & 765 & 1114 &
\begin{tabular}[t]{@{}l@{}} $D_{i\in\{3,8,15,17\!-\!18,34\!-\!35,39,46,50\}}$ \\ \end{tabular} \\
$L_{20}$: Rajasthani     & 694 & 36 & \cellcolor{midgreen}755 & 694 & 62 & \cellcolor{midred} 0 & 694 & 36 & \cellcolor{midgreen}755 &
\begin{tabular}[t]{@{}l@{}} $D_{i\in\{15,28,39\}}$ \\ \end{tabular} \\
$L_{21}$: Sanskrit       & 1321 & 1243 & \cellcolor{midgreen}1325 & 133 & 999 & \cellcolor{midred} 0 & 1321 & 1243 & \cellcolor{midgreen}1325 &
\begin{tabular}[t]{@{}l@{}} $D_{i\in\{15,17,18,33\!-\!35,46\}}$ \\ \end{tabular} \\
$L_{22}$: Santali        & 240 & 240 & \cellcolor{midgreen}245 & \cellcolor{midgreen}245 & \cellcolor{midred}0 & \cellcolor{midred}0 & 240 & 240 & \cellcolor{midgreen}245 &
\begin{tabular}[t]{@{}l@{}} $D_{i\in\{18,35,46\}}$ \\ \end{tabular} \\
$L_{23}$: Sindhi         & 64 & 64 & \cellcolor{midgreen}68 & 56 & \cellcolor{midred} 0 & \cellcolor{midred} 0 & 64 & 64 & \cellcolor{midgreen}68 &
\begin{tabular}[t]{@{}l@{}} $D_{i\in\{8,15,18,35,46\}}$ \\ \end{tabular} \\
$L_{24}$: Tamil          & 1630 & 1629 & \cellcolor{midgreen}2178 & 762 & 1122 & \cellcolor{midred} 29 & 1630 & 1629 & 2156 &
\begin{tabular}[t]{@{}l@{}} $D_{i\in\{3,8,15,17,18,22,23,25,27,29\}}$ \\ \end{tabular} \\
         &  &  &  &  & & &  &  & & 
\begin{tabular}[t]{@{}l@{}} $D_{i\in\{34,35,38,39,45,46,50\}}$ \\ \end{tabular} \\
$L_{25}$: Telugu         & 1389 & 1389 & \cellcolor{midgreen}2005 & 711 & 966 & \cellcolor{midred} 0 & 1389 & 1389 & 2000 & 
\begin{tabular}[t]{@{}l@{}} $D_{i\in\{3,8,15,17,18,22,23,25,27,34\!-\!36,39,46\}}$ \\ \end{tabular} \\
$L_{26}$: Urdu           & 519 & 519 & \cellcolor{midgreen}933 & 207 & 421 & \cellcolor{midred} 0 & 519 & 519 & 873 &
\begin{tabular}[t]{@{}l@{}} $D_{i\in\{3,8,10,17\!-\!18,32,34\!-\!35,39,46,50\}}$ \\ \end{tabular} \\
\bottomrule
\end{tabular}
\caption{Language-wise Duration Distribution with Dataset Mapping (hours). The tasks with the highest and lowest durations for each language are highlighted in green and red, respectively.}
\label{tab:language_task_durations}
\vspace{-1em}
\end{table*}
\textbf{Observations:} Figure~\ref{fig:language_duration_line} shows the total language duration across datasets. Languages such as $L_{2}$, $L_{10}$, $L_{14}$, $L_{16}$, $L_{19}$, $L_{21}$, $L_{24}$, and $L_{25}$ have duration of more than 1000 hours. However, several languages such as $L_{3}$, $L_{5}$, $L_{11}$, $L_{12}$, $L_{13}$, $L_{15}$, $L_{22}$, and $L_{23}$ urgently need more data. Furthermore, Table~\ref{tab:language_task_durations}  reveals stark disparities in language coverage across tasks. Languages such as $L_{2}$, $L_{7}$, $L_{8}$, $L_{10}$, $L_{14}$, $L_{16}$, $L_{17}$, $L_{18}$, $L_{19}$, $L_{20}$, $L_{21}$, $L_{24}$, $L_{25}$, and $L_{26}$ dominate, with at least 500 hours of speech for most tasks. These languages benefit from extensive corpora that yield hundreds of hours of usable audio. On the other hand, there is an urgent need of speech datasets in $L_{3}$, $L_{11}$, $L_{12}$, $L_{13}$, $L_{15}$, $L_{22}$, and $L_{23}$ for all the tasks as current duration is less than 500 hours for each of these tasks. These patterns underscore that task-critical data is concentrated in a handful of languages, leaving many Indian languages under-served. 

Table~\ref{tab:language_task_durations} reveals a highly imbalanced distribution of available speech data across tasks. Across tasks, \(T_{3}\) (LID) and \(T_{9}\) (GRE) frequently exceeds \(T_{1}\) (MO-ASR/STT) and \(T_{2}\) (ML-ASR/STT), indicating that multilingual pooling naturally provides LID and GRE ready data even when transcribed content is sparse. In contrast, \(T_{6}\) (SER) remains the least populated axis, while \(T_{1}\), \(T_{2}\), \(T_{7}\) (MO-TTS), and \(T_{8}\) (ML-TTS) collectively accumulate the majority of hours. Identical durations across \(T_{1}\), \(T_{2}\), \(T_{7}\), and \(T_{8}\) for several languages suggest shared corpora rather than distinct recordings.

Coverage under \(T_{5}\) (ADD) is absent for \(L_{3}\), \(L_{11}\), \(L_{12}\), \(L_{17}\), \(L_{22}\), and \(L_{23}\); limited for \(L_{4}\), \(L_{5}\), \(L_{9}\), \(L_{13}\), and \(L_{20}\); but comparatively higher for \(L_{2}\), \(L_{8}\), \(L_{10}\), \(L_{14}\), \(L_{16}\), \(L_{19}\), \(L_{21}\), \(L_{24}\), and \(L_{25}\). Dravidian languages such as \(L_{10}\), \(L_{14}\), \(L_{24}\), and \(L_{25}\) exhibit strong presence in \(T_{3}\), \(T_{5}\), and \(T_{9}\), reflecting diverse annotations and sustained collection efforts relative to many Indo-Aryan counterparts.  

Language-specific gaps remain evident. \(L_{3}\) shows about 120 hours across tasks yet none in \(T_{4}\) (SV/SID), \(T_{5}\) and \(T_{6}\). A recurring observation is the dominance of shared datasets, particularly \(D_{15}\), \(D_{18}\), \(D_{39}\), and \(D_{46}\), which populate multiple cells in the matrix and span \(L_{1}\), \(L_{4}\), \(L_{5}\), \(L_{8}\), \(L_{13}\), \(L_{15}\), \(L_{16}\), \(L_{18}\), \(L_{19}\), \(L_{24}\), and \(L_{25}\), across \(T_{1}\), \(T_{3}\), \(T_{7}\), and \(T_{8}\), demonstrating their multitask importance within the current resource landscape.

\section{Conclusion and Future Work}
This paper introduced Task-Lens, a cross-task utility oriented lens for profiling speech datasets. Task-Lens involves four stages: dataset discovery, dataset filtering, feature extraction, and utility mapping. We conducted a comprehensive cross-task profiling of 50 Indian speech datasets spanning 26 languages and 91,257 hours of audio to assess their readiness for nine speech tasks based on the available metadata. Next, through Task-Lens outputs, we answered four vital research questions: (1) Which speech tasks does each
dataset currently support? (2) What enhancements could improve cross-task applicability? (3) Which tasks lack sufficient data support? (4) Which Indian languages have adequate per-task coverage, and where do gaps persist? It turns out that datasets BhasaAnuvaad \citep{bhashanuvaad}, South Asian Crowdsourced Speech \citep{kjartansson-etal-sltu2018}, IndicSUPERB \citep{javed2023indicsuperb}, IndicVoices-R \citep{indicvoices_r}, MS Indic Speech \citep{msindicspeech2024}, Nexdata AI 759h \citep{nexdata2025hindi}, NPTEL2020 \citep{nptel2020}, Rasa \citep{rasa_dataset}, IndicSynth \citep{indicsynth}, NIRANTAR \citep{nirantar2025interspeech}, EmoTa: A Tamil Emotional Speech Dataset \citep{thevakumar2025emota}, IndicFake \citep{anonymous2025indicfake}, BanglaSER \citep{das2022bangalser}, SUBESCO \citep{sultana2021subesco}, BanSpEmo \citep{sultana2025banspemo}, SEA\_Spoof \citep{wu2025seaspoofbridginggapmultilingual}, IndieFake \citep{kumar2025indiefakedatasetbenchmarkdataset}, Bangla Speech Corpus \citep{ahmed2020banglacorpus}, and Indic-TEDST \citep{sethiya2024indictedst} lead the rankings. In contrast, many datasets need richer metadata diversity and greater scale. Similarly, language identification and gender recognition are moderately resourced. However, speaker verification / identification, audio deepfake detection, and emotion recognition remain critically underserved. From a language perspective, Bhojpuri, Dogri, Kashmiri, Konkani, Maithili, Manipuri, Santali, and Sindhi urgently need more data, as existing resources contain under 400 hours of speech.


This work opens several avenues for future research. By surfacing cross-task utility and highlighting dataset gaps, Task-Lens can give researchers immediate guidance on which dataset to use for their research problem, enabling them to focus on model innovation rather than time-consuming dataset curation. Furthermore, Task-Lens addresses inefficient use of existing resources and limited awareness of task appropriate datasets in low resource languages. Generic AI tools and raw web searches list datasets based on their original intent. However, they rarely provide reliable cross-task profiles for efficient use. Additionally, by highlighting under-resourced languages and tasks, this work supports researchers interested in creating datasets for under-served areas. With over 7,000 languages spoken worldwide, extending Task-Lens can help advance inclusive speech processing research across diverse languages and tasks.

\section{Limitations}
Despite the breadth and utility of Task-Lens, we acknowledge certain limitations:

\begin{enumerate}
  \item \textbf{Language and Task Coverage:} Our exploration is limited to nine core speech tasks, 26 languages and 50 Indian speech datasets, excluding less accessible datasets and emerging tasks (e.g., code-switching ASR). However, we plan to integrate additional community-contributed resources to broaden language and task support. Furthermore, all per-dataset feature values will be released upon publication to facilitate Task-Lens extension.

  \item \textbf{Quality-aware utility estimation:} The current readiness check treats feature presence as binary. Incorporating signal quality, annotation quality, and basic sensitivity analyses can calibrate utility scores as well and strengthen their connection to downstream performance.
\end{enumerate}

Despite these limitations, the NLP community will gain a practical, transparent foundation for efficient dataset selection, a clear roadmap for extending exploration to new languages, tasks, and standards, and practical insights into data resource gaps that can drive more resource contributions in the future\footnote{We used Grammarly and ChatGPT to assist with sentence construction and language refinement. While submitting our paper, we indicated it as a 20\% use of generative AI.}.


\section{Ethical Considerations}
We encourage adoption of Task-Lens to support systematic cross-task profiling across diverse languages and tasks. Most datasets in our survey carry Creative Commons Attribution (CC BY 4.0) licenses; a smaller subset uses commercial terms or research-/academic-use-only terms (e.g., ELRA research, LDC-IL), alongside a few custom or paid licenses. All of these permit non-commercial cross-task research. One dataset \citep{wu2025seaspoofbridginggapmultilingual} is released under CC BY-NC-ND 4.0, which restricts derivative uses; we nevertheless include it due to its uniquely valuable synthetic speech data for Southeast Asian languages. We will release verified license details for all included datasets upon publication to promote clarity and responsible reuse. Researchers should confirm each dataset’s license and README, respect usage restrictions, and ensure compliance with privacy, consent, and data-protection norms.

\section{Bibliographical References}\label{sec:reference}

\bibliographystyle{lrec2026-natbib}
\bibliography{lrec2026-example}


\end{document}